\documentclass[journal]{IEEEtran}
\IEEEoverridecommandlockouts                              

\usepackage{graphics} 
\usepackage{epsfig} 
\usepackage{amsmath} 
\usepackage{color}
\usepackage{multirow}
\usepackage[breaklinks,colorlinks,linkcolor=black,citecolor=black,urlcolor=black]{hyperref}
\usepackage{diagbox}
\usepackage{cite}
\usepackage{caption}

\begin{document}

\title{\LARGE \bf
	Cross Scene Prediction via Modeling Dynamic Correlation using Latent Space Shared Auto-Encoders
}
\author{Shaochi~Hu, Donghao~Xu,~\IEEEmembership{Member,~IEEE,}
	and Huijing~Zhao, ~\IEEEmembership{Member,~IEEE,}
	\thanks{This work is supported by the National Natural Science Foundation of China (61973004).}
	\thanks{S.Hu, D.Xu and H.Zhao are with the Peking University, with
		the Key Laboratory of Machine Perception (MOE), and also with the School of
		Electronics Engineering and Computer Science}
	\thanks{Correspondence: H. Zhao, zhaohj@cis.pku.edu.cn.}
}

\let\oldtwocolumn\twocolumn
\renewcommand\twocolumn[1][]{%
	\oldtwocolumn[{#1}{
		\begin{center}
			\includegraphics[keepaspectratio=true,width=0.9\textwidth]{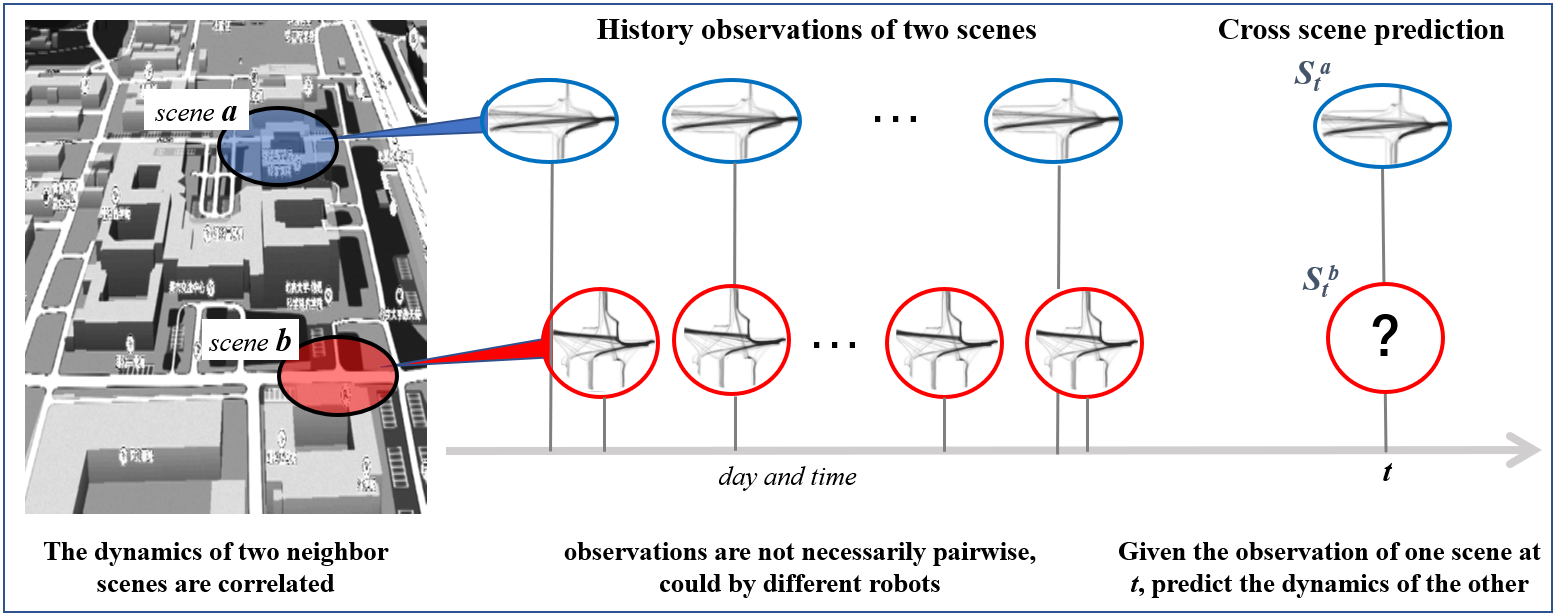}
			\captionof{figure}{Can we cross-scene prediction via modeling the correlations of scene dynamics on unsynchronized history observations?}
			\label{fig:overview}
		\end{center}
	}]
}
\maketitle
\thispagestyle{empty}
\pagestyle{empty}

\begin{abstract}
	This work addresses on the following problem: given a set of unsynchronized history observations of two scenes that are correlative on their dynamic changes, the purpose is to learn a cross-scene predictor, so that with the observation of one scene, a robot can onlinely predict the dynamic state of another. A method is proposed to solve the problem via modeling dynamic correlation using latent space shared auto-encoders. Assuming that the inherent correlation of scene dynamics can be represented by shared latent space, where a common latent state is reached if the observations of both scenes are at an approximate time, a learning model is developed by connecting two auto-encoders through the latent space, and a prediction model is built by concatenating the encoder of the input scene with the decoder of the target one. Simulation datasets are generated imitating the dynamic flows at two adjacent gates of a campus, where the dynamic changes are triggered by a common working and teaching schedule. Similar scenarios can also be found at successive intersections on a single road, gates of a subway station, etc. Accuracy of cross-scene prediction is examined at various conditions of scene correlation and pairwise observations. Potentials of the proposed method are demonstrated by comparing with conventional end-to-end methods and linear predictions.
\end{abstract}

\section{INTRODUCTION}
The ability to deal with dynamic change of environment is important for robots to achieve lifelong and robust autonomy. Map-based localization approaches could fail if the map is far different from the current environment, and planning can be harder without the knowledge of the dynamic environment. 

Some methods are proposed to model the dynamic change of the environment in different aspects. The basic idea is maintaining a database that saves all different observations of the environment, and localization is performed in all past map\cite{Churchill2013a}. However, this is only a kind of data collection without the modeling process for the change of environment. And with the increasing of the database, computation efficiency and localization in real-time are rapidly influenced. In some scenarios, the change of environment is periodic, which inspires frequency map approach that models the dynamic environment as the sum of some periodic functions\cite{Krajnik2017}. This is a signal level modeling and is able to predict the future of the environment. In general, the dynamic change between neighboring scenes are related, such as traffic flow changing between intersections, and learning the relationships between them is another kind of modeling. Nicholas\cite{Carlevaris-bianco} applies mutual information based method to learn the temporal observability relationships between them.

This work addresses on a new problem: can we make inference by modeling the correlations of scene dynamics on history observations? As illustrated in Fig. 1, two scenes are adjacent, such as nearby gates of the campus, successive intersections on a single road, gates of a subway station, etc. Dynamic changes of the two scenes are correlated, which are triggered by some common events, such as working and teaching schedules of the campus, the period of the traffic signal, the train's pit stop and so forth.
There have been history observations of both scenes, whereas they are not synchronized as they could be measured by different robots, i.e. the observations are not necessarily in pairs.
At a certain time, given the observation of one scene, we want to predict the dynamic state of the others.
This research proposes a method of cross-scene prediction via modeling dynamic correlation using latent space shared auto-encoders,
which is developed based on an assumption that the inherent correlation of scene dynamics can be represented by shared latent space, and
a common latent state is reached if the observations of both scenes are at an approximate time.
A learning model is thus developed by connecting two auto-encoders through the latent space, and a prediction model is built by concatenating the encoder of the input scene with the decoder of the target one.
Simulation datasets are generated, where two scenes are designed imitating the dynamic flows at two adjacent gates of Peking Univ., and a simulator is developed to obtain scene maps for hours. Accuracy of cross-scene prediction is examined, and the performance at various conditions of scene correlation and pairwise observations are elaborated. Potentials of the proposed method are demonstrated by comparing with conventional end-to-end methods and linear predictions.

This paper is organized as follows. The related works are reviewed in Section\ref{section:relatedWorks}. Section\ref{section:methodology} explain the detail of our method. Section\ref{section:implementationDetails} and Section\ref{section:simulation} show implementation details of model and simulation. Experimental results are in Section\ref{section:experiments}.

\section{RELATED WORKS}\label{section:relatedWorks}
There are some researches focus on how to model or predict the dynamic change of the environment. Spectrum-analysis based methods\cite{Krajnik2014a}\cite{Krajnik2017} discretize environment into binary voxels indicating they are occupied or not, and model each voxel as the sum of a series of periodic signals by the frequency spectra of observed data. Their ability to predict environment improve localization accuracy\cite{Krajnik2014c}\cite{Fentanes2015a} and efficiency of map updating\cite{Santos2016b}\cite{Duckett}. Some methods apply long-term and short-term memory in dynamic scene mapping to remove nonexistent features and increase emerging features\cite{Dayoub2008b}\cite{Morris2014}\cite{Berrio2019}. The bag-of-word method is also used to predict image between seasons\cite{Neubert2013}. Mutual information based method\cite{Carlevaris-bianco} predict images of neighboring scenes by calculating the correlation of collected data, this work is similar to ours but the essence is different because it only learns the temporal relationship between data without considering what makes data correlate and model the essence.


In this paper, the dynamic of a scene is caused by moving objects like pedestrians, and there are lots of studies about traffic behavior and scene modeling in the surveillance field. There is a shift from detecting-and-tracking of vehicle state and defining interested events towards machine learning-based approaches to automatically extract meaningful pattern\cite{Morris2013}. Similar trajectories are clustered to model structure or path of scene\cite{Makris2005}\cite{Piciarelli2006}. Topic model based methods convert conceptions of natural language processing into traffic behavior, and LDA\cite{Kwak2011}\cite{Song2011}/HDA\cite{Wang2009}\cite{Haines2011} approaches achieve good results in scene modeling without accurate tracking. Scene modeling methods in the surveillance field are mainly used for abnormal events detection or scene semantic understanding\cite{Saleemi2009}, but there are few predictions for the future of the full scene. Besides, they do not consider the correlation between neighboring scenes. We can learn some methods in this field, but our conception of scene modeling is essentially different from theirs.

This work makes an attempt to model the dynamic correlation between neighboring scenes on simulation datasets. In order to quantify how the correlation influences our algorithm, we generate datasets with different correlation coefficient between scenes. Training data with different pairwise observations are randomly sampled to simulate robots data acquisition situation in the true world.

\section{METHODOLOGY}\label{section:methodology}

\subsection{Problem definition}

As illustrated in Fig.1, $a$ and $b$ are two neighbor scenes such as adjacent gates of a campus or consecutive intersections on a single road, where the scene dynamics are strongly correlated.
Let $\mathbf{S}^a=\{<S_1^a,t_1^a>,...,<S_{n}^a,t_{n}^a>\}$ and $\mathbf{S}^b=\{<S_1^b,t_1^b>,...,<S_{m}^b,t_{m}^b>\}$ be the history observations of both scenes.
$S_i^k$ denotes the $i$th observation of scene $k$ at time $t_i^k$, which can be a grid map that represents the dynamic state of the scene.
The observations of both scenes are not necessarily pairwise in time, i.e. $\{t_1^a,...,t_{n}^a\} \neq \{t_1^b,...,t_{n}^b\}$, as they could be obtained independently by different robots.

The purpose of this work is to learn a predictor ${\cal F}$ on $\mathbf{S}^a$ and $\mathbf{S}^b$ by addressing the correlation of scene dynamics, where given the observation $S_i^j$ of one scene at the current time $t$, predict the dynamic state of the other, e.g.
\begin{equation}
\hat{S}.^b,t = \mathbf{F} (S.^a,t)
\end{equation}
\begin{equation}
\hat{S}.^a,t = \mathbf{F} (S.^b,t)
\end{equation}

The formulations can be easily extended to define the problems involving three or more scenes.

\subsection{Modeling dynamic correlation using latent space shared auto-encoders}

Assumes that there exists a latent space $\mathbf{Z}$ that records the inherent correlation of scene dynamics at $a$ and $b$,
after encoding the observations $\mathbf{S}^a$ and $\mathbf{S}^b$ individually to the latent space $\mathbf{Z}$,
\begin{equation}
Z_i^a = \mathbf{E}_a (S_i^a)
\end{equation}
\begin{equation}
Z_j^b = \mathbf{E}_b (S_j^b)
\end{equation}
$S_i^a$ and $S_j^b$ may share a common state, i.e.
\begin{equation}
\Delta Z = ||Z_i^a - Z_j^b||_2 \rightarrow 0 \nonumber
\end{equation}
if they are the observations of an approximate time, i.e.
\begin{equation}
\Delta t = dis (t_i^a , t_j^b) \rightarrow 0 \nonumber
\end{equation}
where $dis$ is an operator of time difference by addressing the periodic nature of scene dynamics.

\begin{figure}[tp]
	\centering
	\includegraphics[keepaspectratio=true,width=\linewidth]{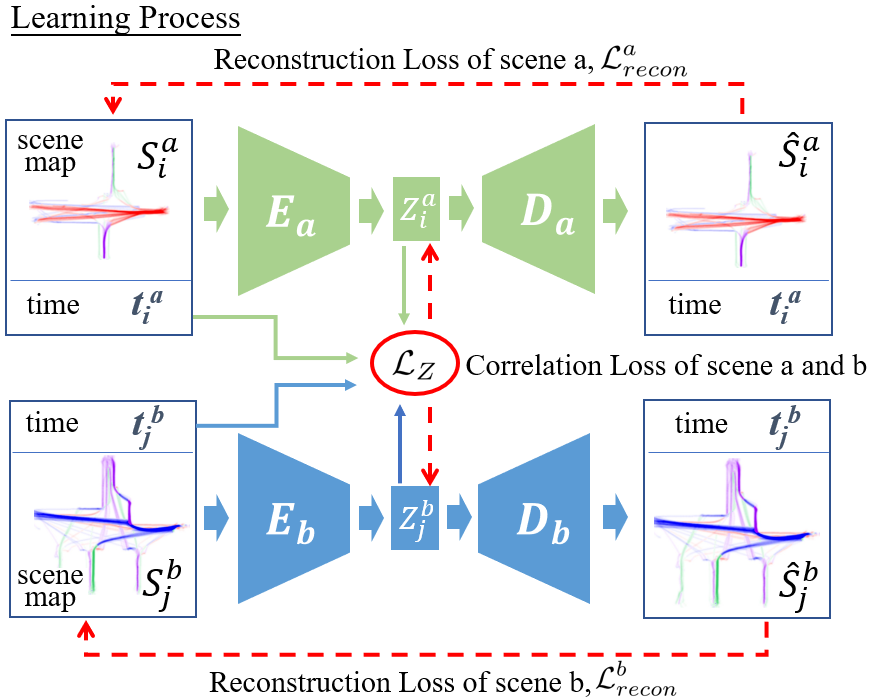}
	\caption{Modeling dynamic correlation using latent space shared auto-encoders.}
	\label{networkArchitecture}
\end{figure}

As illustrated in Fig.\ref{networkArchitecture}, the procedure is modeled by combining the auto-encoder structures in this work.
Given a pair of history observations of both scenes $S_i^a$ and $S_j^b$ that are measured at $t_i^a$ and $t_j^b$ respectively,
each scene map is processed individually through the corresponding encoding-decoding path of the scene.
\begin{eqnarray}
Z_i^a = \mathbf{E}_a (S_i^a),
\hat{S}_i^a = \mathbf{D}_a (Z_i^a)\\
Z_j^b = \mathbf{E}_b (S_j^b),
\hat{S}_j^b = \mathbf{D}_b (Z_j^b)
\end{eqnarray}

Two reconstruction losses $\mathcal{L}_{recon}^a$ and $\mathcal{L}_{recon}^b$ are defined to evaluate the auto-encoder's accuracy of each scene,
\begin{eqnarray}
\mathcal{L}_{recon}^a = \left\|S_{i}^{a} - \hat{S}_{i}^{a}\right\|_2\\
\mathcal{L}_{recon}^b = \left\|S_{j}^{b} - \hat{S}_{j}^{b}\right\|_2
\end{eqnarray}
and a correlation loss are defined to constrain equivalent latent states if the scene dynamics are observed at approximative time points.
\begin{eqnarray}
&&\mathcal{L}_Z = \exp(-c\cdot \Delta t) \cdot \left\| Z_{i}^{a} - Z_{j}^{b}   \right\|_2\\
&&\Delta t = dis (t_i^a , t_j^b) \nonumber
\end{eqnarray}
Therefore, model learning is conducted by optimizing the following total loss
\begin{equation}
\mathop{\min}_{E_{a},E_{b},D_{a},D_{b}}  \mathcal{L}_{recon}^a + \mathcal{L}_{recon}^b + \lambda\mathcal{L}_{Z}
\label{optTarget}
\end{equation}
where $\lambda$ is a hyperparameter that is assigned 0.1 in this research.

\begin{figure}[tp]
	\centering
	\includegraphics[keepaspectratio=true,width=\linewidth]{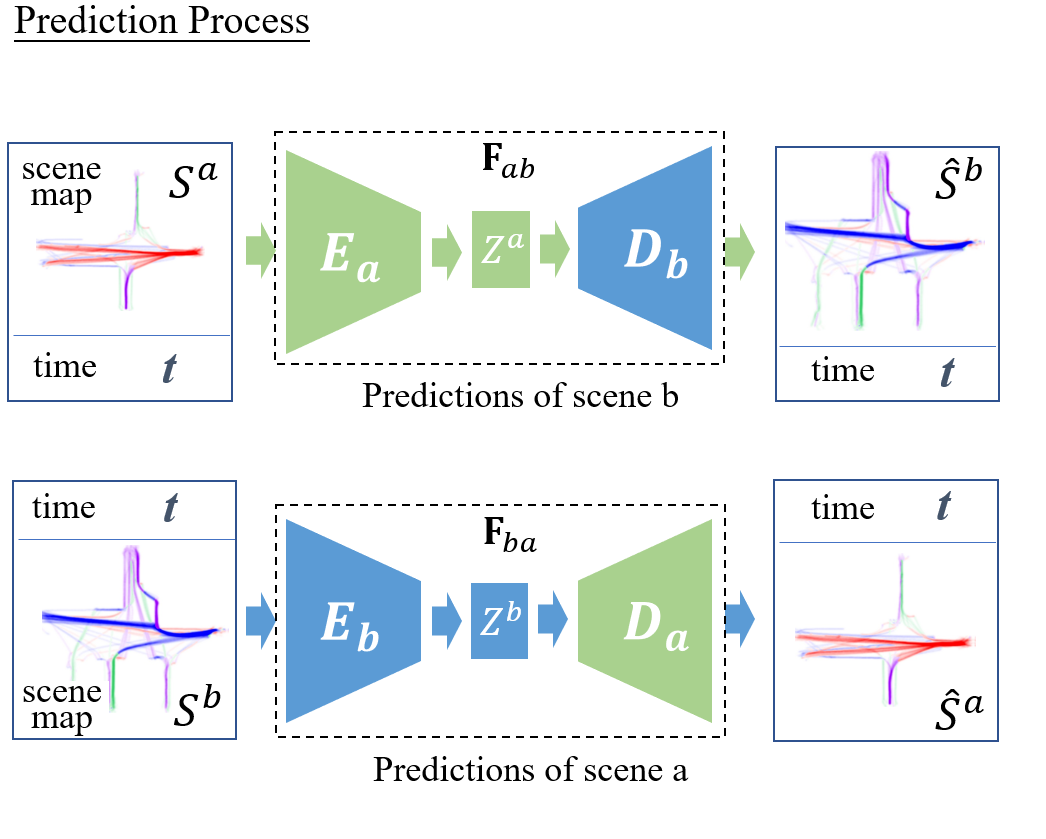}
	\caption{Cross-scene prediction by concatenating the encoder of the input scene with the decoder of the target one.}
	\label{predictionProcess}
\end{figure}

The prediction model $\mathbf{F}$ is built by concatenating the encoder of the input scene with the decoder of the target one, as illustrated in Fig.\ref{predictionProcess}.
For example, at the current time $t$, given the observation $S^a$ of scene $a$, the dynamic state of scene $b$ can be predicted by
\begin{eqnarray}
&&\hat{S}^b,t = \mathbf{F}_{ab} (S^a,t) \\
&&\mathbf{F}_{ab}=E_{a}oD_{b}
\end{eqnarray}
and vice versa
\begin{eqnarray}
&&\hat{S}^a,t = \mathbf{F}_{ba} (S^b,t) \\
&&\mathbf{F}_{ba}=E_{b}oD_{a}
\end{eqnarray}


\section{IMPLEMENTATION DETAILS}\label{section:implementationDetails}
\subsection{Scene map}
A grid map is used to represent the dynamic state of a scene, where each pixel is a four-dimensional vector, recording the number of dynamic objects passing through the location during a short time window $\tau$ on four discretized directions.
In this research, the map has a dimension of $512 \times 512$ and a pixel size of 0.2 meters, $\tau=1$ min, and the four directions correspond to the East, West, South, and North in the world coordinate system.
In this paper, pixels of scene map are visualized by the most dominant flow crossing the pixels at the time, where red, blue, purple and green represents the four discretized directions to the west, east, south and north, respectively, the brighter the color, the higher the dynamic flow.

\subsection{Network design}
\begin{figure}[hb]
	\centering
	\includegraphics[keepaspectratio=true,width=\linewidth]{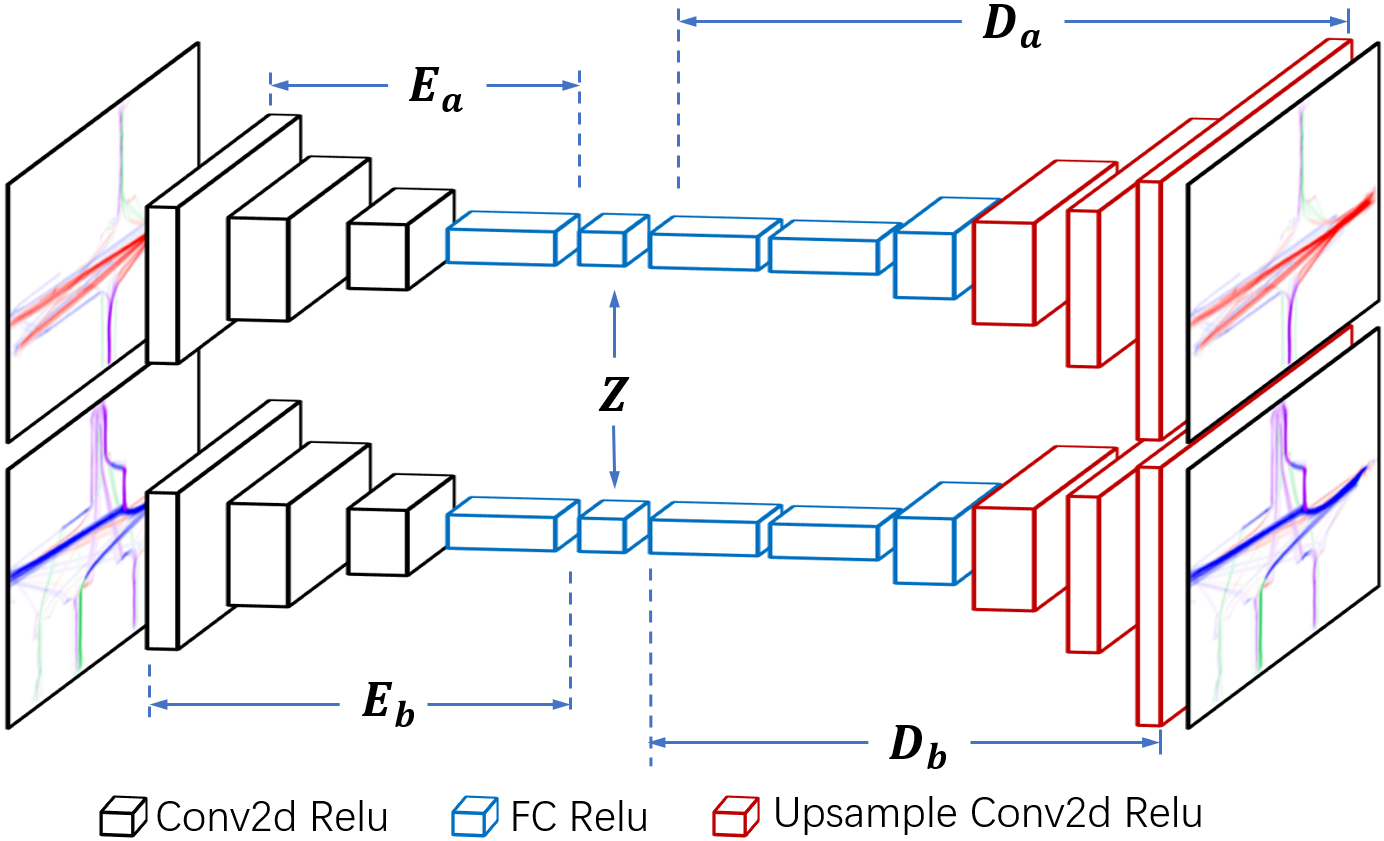}
	\caption{The network struture of autoencoders. }
	\label{networkArch}
\end{figure}

As illustrated in Fig.\ref{networkArch}, the network contains two autoencoders that have the same structure.
We use PyTorch framework to realize the autoencoder \cite{Chakravarty2019
}, which is composed of convolutional, fully connected and upsample layers.

There is no pooling layer in the encoder part, and input size is reduced only by convolutional layers with stride=2. For the decoder part, we use $\times 2$ upsampling with same-padding convolutional layers to extend the size of the input, instead of deconvolutional layers. Such a structure can make the network retain more information.

In encoders, the input size changes from $512 \times 512 \times 4$ to $64 \times 64 \times 8$ by 3 Conv2d layes, and then reduced to 2 dimensions(the latend variable Z) by 2 FC layers. In decoders, the size is extented from 2 to $64 \times 64 \times 8$ by 3 FC layes, and then restores to $512 \times 512 \times 4$ by 3 Upsample and Conv2d layers.

\section{SIMULATION DATASETS}\label{section:simulation}

\begin{figure}[]
	\centering
	\includegraphics[keepaspectratio=true,width=\linewidth]{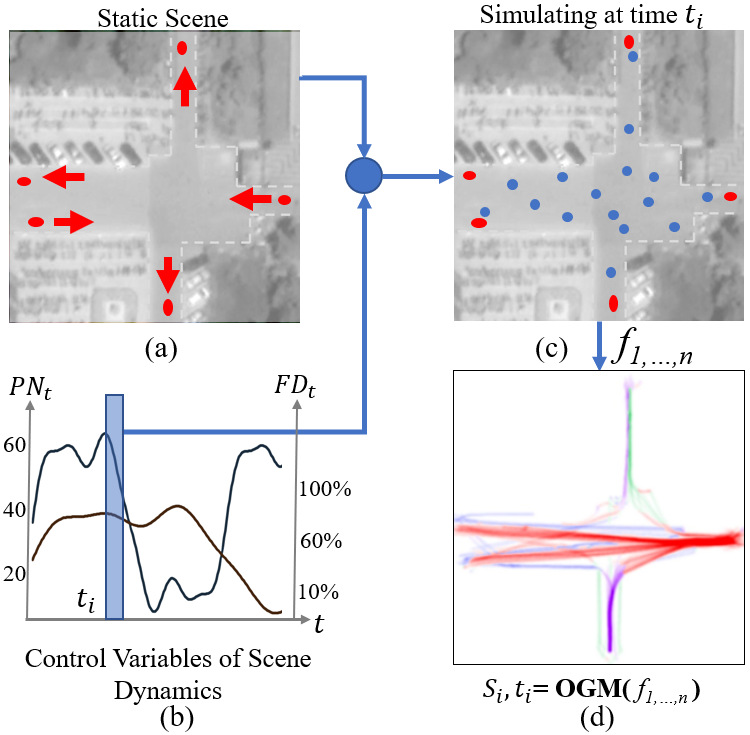}
	\caption{Simulation pipeline. (a)scene layout, (b)series of control variables of the dynamic flows, (c) a simulation frame of the dynamic objects(blue points), (d) a scene map on a frame sequence during a short time window.}
	\label{simulatorPipeline}
\end{figure}

A simulator is developed to generate simulation datasets for experiments.
The simulation pipeline is shown in Fig.\ref{simulatorPipeline}. 
Without loss of generality, we assume that each scene has its inherent structure of the dynamic flows that connect a set of entrance and exit points of the scene, shown as red points in Fig.\ref{simulatorPipeline}(a), whereas the volume of each flow may change with time due to some underlying events.Therefore, time series of a set of control variables are designed as illustrated in Fig.\ref{simulatorPipeline}(b) to guide the simulation of dynamic objects.
In this research, two control variables are designed, which are the total people number $PN_t$ and the main flow direction $FD_t$.
Two main flow directions are defined, where $FD_t$ is the percentage of people entering the campus, leaving the rest $1-FD_t$ going out.
At a time $t$, if the total people number at the frame is less than $PN_t$, new people are generated to meet the insufficient number.
Among the new people, $FD_t$ are generated at the entrance point of the gate following a randomly chosen flow entering the campus, while $1-FD_t$ are generated randomly at the start point of a flow going out of the campus.

People flows are simulated by referring to Helbing's work\cite{Helbing1995}.Each scene map is estimated on a sequence of simulation frames as
\begin{equation}
S_i, t_i = \mathbf{OGM} (f_{1,...,n})
\end{equation}

\begin{figure}[]
	\centering
	\includegraphics[keepaspectratio=true,width=\linewidth]{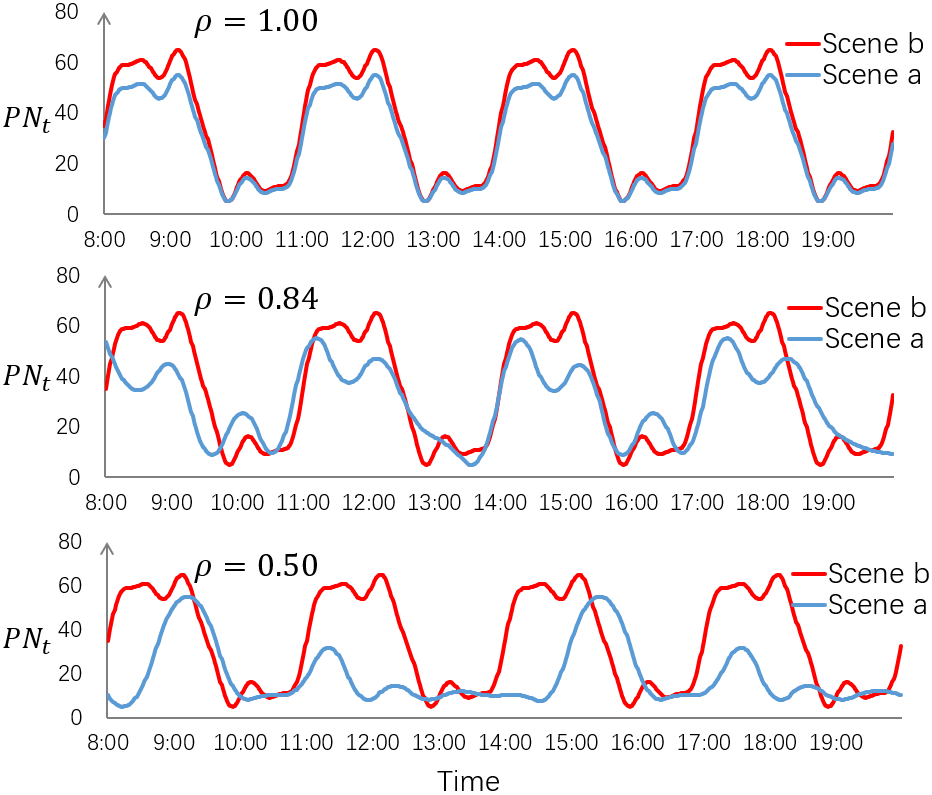}
	\caption{Scene dynamic correlation simulated by designing correlative time series of control variables. Three patterns are designed with different correlation coefficient $\rho$ of the time series on the control variables $PN_t$.}
	\label{peopleNumCurves}
\end{figure}

In this research, simulation frames $f_{1,...,n}$ are generated at 10Hz. Each scene map represent the dynamic state during a short time window of $\tau=1$ min, therefore $n=600$ frames are used to estimate a $S_i$ at $t_i$. 

Two scenes are simulated by imitating the dynamic flows at two adjacent gates of Peking Univ., which are triggered by almost the same events, e.g. working and teaching schedules of the campus.
Similar scenarios can also be found at such as adjacent intersections on a single road, subway stations, gates of a stadium, etc.
Therefore, correlated time series of control variables at both scenes are designed as shown in Fig. \ref{peopleNumCurves}.
Three kinds of patterns are designed with the correlations coefficients $\rho=$1.0, 0.84 and 0.5 of $PN_t$ of two scenes, representing the strong, middle and less correlative scenes.
Here people number $PN_t$ of two scenes are designed to control the correlation of two scenes, and we keep the main flow direction $FD_t$ of two scenes the same.

\begin{figure}[]
	\centering
	\includegraphics[keepaspectratio=true,width=\linewidth]{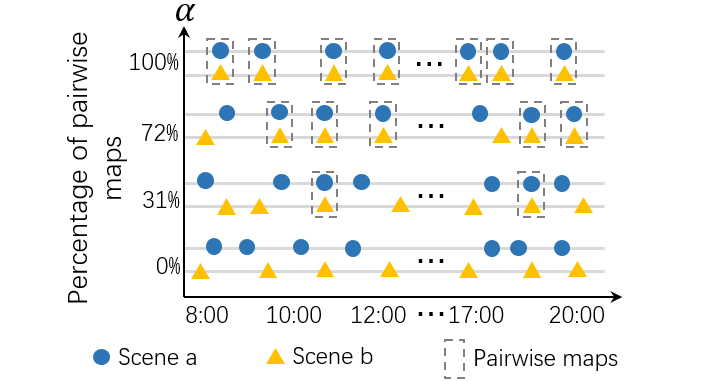}
	\caption{Datasets generation of various percentage of pairwise scene maps, $\alpha=$0\%,31\%,72\%,100\%.}
	\label{fig:alphaDataCollection}
\end{figure}

Following each pattern of time series in Fig.\ref{peopleNumCurves}, a simulation is conducted from 8:00 to 20:00, where 720 scene maps are generated every 1 minute for both scenes.
Part of scene maps are selected to simulate the different data acquisition situation, which have $\alpha=$0\%,31\%,72\%,100\% of pairwise observations as shown in Fig.\ref{fig:alphaDataCollection}. 
Therefore, a total of 12 datasets containing three correlation patterns and four percentages of pairwise observations are generated, which are used in the experiments.
In each particular experiment, although the proposed and baseline methods are trained and test on the same dataset, the number of scene maps used in training could be different due to the requirements on pairwise observations of each method, which is detailed in Tab. \ref{tab:datasets}.



\begin{table}[h]
	\centering
	\caption{The number of maps in training and testing of the cross-scene prediction models}
	\label{tab:datasets}
	\begin{tabular}{|c|c|c|c|c|c|}
		\hline
		\multicolumn{2}{|c|}{\begin{tabular}[c]{@{}c@{}}\diagbox{Datasets}{Methods}\end{tabular}}                  & $\mathbf{F}_{our}$ & $\mathbf{F}_{E2E}$ & $\mathbf{F}_{E2E\Delta t}$ & $\mathbf{F}_{linear}$ \\ \hline
		\multirow{4}{*}{Training} & $\alpha=0$     & 72                 & 0                  & 72                         & -                     \\ \cline{2-6} 
		& $\alpha=31\%$  & 72                 & 22                 & 72                         & -                     \\ \cline{2-6} 
		& $\alpha=72\%$  & 72                 & 51                 & 72                         & -                     \\ \cline{2-6} 
		& $\alpha=100\%$ & 72                 & 72                 & 72                         & -                     \\ \hline
		Testing                   & -              & 36                 & 36                 & 36                         & 36                    \\ \hline
	\end{tabular}
\end{table}

\section{EXPERIMENTAL RESULTS}\label{section:experiments}

\subsection{Evaluation measures}

\subsubsection{Prediction error}
Given two maps $S_1$ and $S_2$ of size $W\times H\times C$, mean square error(MSE) is used to measure the difference between them
\begin{equation}
{\cal D}_s(S_1,S_2) = \frac{1}{W\times H\times C}\sum_{W,H,C}^{}(S_1-S_2) ^2
\end{equation}
Subsequently, for a predicted map $\hat{S}$ with a ground truth $S$, the prediction error ${\cal E}_{s}$ is defined as
\begin{equation}
{\cal E}_{s}(\hat{S})={\cal D}_s(\hat{S},S)
\end{equation}

\subsubsection{Dataset variance}
A scene map describes the dynamic state of a scene, which is generated by taking statistics on the data frames during a short time window around the time, i.e. $n_f=600$ frames during $\tau=1$ min in this research. A scene map has the nature of randomness due to uncontrollable scene dynamics and the method of time windowing, the variance of such randomness is an important reference to prediction accuracy.

Given each series $\cal C$ of control variables, simulations are conducted for $n$ times.
At each sampled time $t$ corresponding to frame number $i_t$, a time window $[i_0,i_0+n_f]$ is randomly chosen for $m$ times with $i_0 \in [i_t-n_f,i_t]$, and a scene map is subsequently generated on data frames $f_{i0,...,i_0+n_f}$. Therefore, $n*m$ scene maps $\{S_1,...S_{n*m}\}$ are generated, and inherent variance of scene map for $\cal C$ and $t$ is estimated below.
\begin{equation}
{\cal V}_s ({\cal C},t)= \frac{1}{n*m}\sum_{i=1}^{n*m}{\cal D}_s(S_i,\overline{S})
\end{equation}
where, $\overline{S} = \frac{1}{n*m}\sum_{i=1}^{n*m}S_i$ is the mean map.

By repeating the above estimations at all sampled time points $t \in \Omega_t$ and control series $\cal C \in \Omega_C$, variance at the level of control series and data sets can also be found.
\begin{eqnarray}
{\cal V}_c ({\cal C})= \frac{1}{|\Omega_t|}\sum_{t \in \Omega_t} {\cal V}_s ({\cal C},t) \\
{\cal V}_d = \frac{1}{|\Omega_t \times \Omega_C|}\sum_{t,C \in \Omega_t \times \Omega_C} {\cal V}_s ({\cal C},t)
\end{eqnarray}

Dataset variance is the lower bounder of prediction error for any methods, and the closer the prediction error to the dataset variance, the better the result. 

\subsection{Baseline methods}\label{section:baselineMethods}

\begin{figure}[b]
	\centering
	\includegraphics[keepaspectratio=true,width=\linewidth]{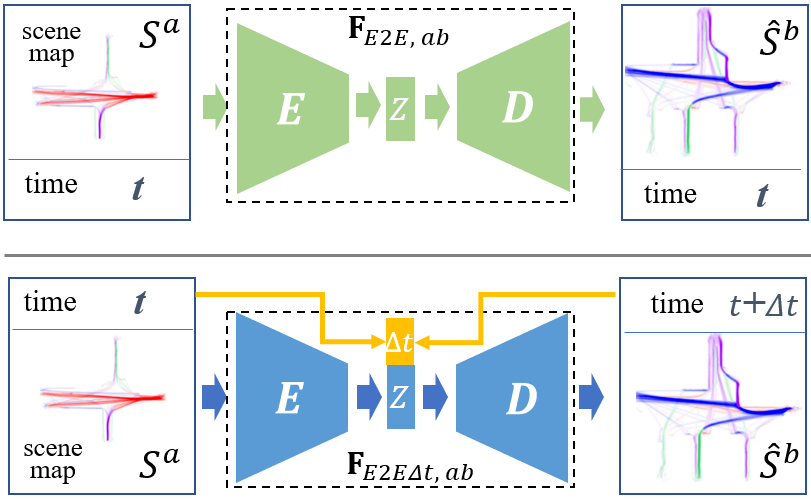}
	\caption{The baseline methods. Top: conventional end-to-end prediction trained by pairwise maps only. Down: end-to-end prediction with compensation of time difference.}
	\label{endtoend}
\end{figure}

\subsubsection{$\mathbf{F}_{E2E}$ - Conventional end-to-end prediction}
By using only pairwise observations in training datasets, a pair of conventional end-to-end predictors $\mathbf{F}_{E2E}$ can be trained as illustrated in Fig. \ref{endtoend} to predict a $\hat{S}_b$ of scene $b$ on $S_a$ of $a$, and vice versa.

\begin{eqnarray}
&&\hat{S}^b, t=\mathbf{F}_{E2E,ab} (S^a,t) \\
&&\hat{S}^a, t=\mathbf{F}_{E2E,ba} (S^b,t)
\end{eqnarray}

\subsubsection{$\mathbf{F}_{E2E\Delta t}$ - End-to-end prediction with compensation of time difference}
However, the observations are not necessarily pairwise, which could be measured by a multi-robot system.
Therefore, the pairwise observations in training datasets are limited when $alpha=$31\%, and none when $alpha=$0\% for $\mathbf{F}_{E2E}$, as shown in TABLE \ref{tab:datasets}.
A pair of conventional end-to-end predictors with compensation of time difference is

\begin{eqnarray}
&&\hat{S}^b, t+\Delta t=\mathbf{F}_{E2E\Delta t,ab} (S^a,t,\Delta t) \\
&&\hat{S}^a, t+\Delta t=\mathbf{F}_{E2E\Delta t,ba} (S^b,t,\Delta t)
\end{eqnarray}

\subsubsection{$\mathbf{F}_{linear}$ - Linear interpolation}
A scene map can also be predicted by finding two history observations of the nearest time by considering the periodic nature of scene dynamics and conducting linear interpolation.
Let $S_.(t_1)$ and $S_.(t_2)$ be the two history observations of the scene at time $t_1$ and $t_2$ respectively, and a predicted one of time $t$ is estimated as below.

\begin{equation}
\hat{S}_., t= \frac{S_.(t_2) - S_.(t_1)}{t_2-t_1} \times (t-t_2) + S_.(t_2)
\end{equation}

\subsection{Prediction results}
\begin{figure*}[h]
	\centering
	\includegraphics[keepaspectratio=true,width=\linewidth]{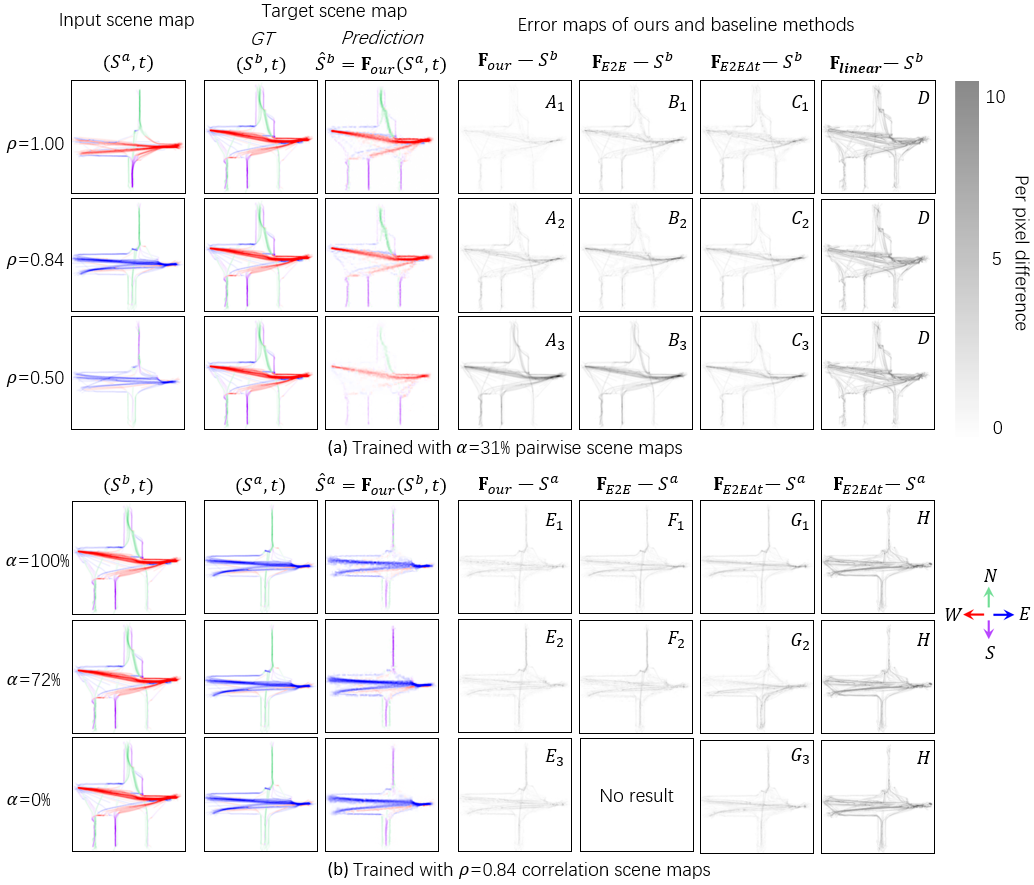}
	\caption{Case study of prediction results, comparison with baseline methods at various conditions of scene correlation ($\rho$) and pairwise observations ($\alpha$).}
	\label{predictionResults}
\end{figure*}

We evaluated our method's performance at various conditions of scene correlation($\rho$) and pairwise observations($\alpha$) comparing with baseline methods, and the quantitative results are shown in TABLE.\ref{predictionAccuracy}. 
Besides, case study of prediction results is illustrated in Fig.\ref{predictionResults}. Given the input scene map, the ground truth map of the other scene is compared with our prediction result, and error maps of ours and baseline methods are also shown on the right four columns.
Finally, the study about per map prediction error on the single dataset is exhibited in Fig.\ref{fig:studyOnADataset}.

\begin{table*}[h]
	\centering
	\caption{Average prediction error on each dataset corresponding to a pair of $\rho$ and $\alpha$}
	\label{predictionAccuracy}
	\begin{tabular}{|c|c|c|c|c|c|c|c|c|c|c|c|c|c|}
		\hline
		\multirow{2}{*}{\begin{tabular}[c]{@{}c@{}}\diagbox[width=7em]{Methods}{Datasets}\end{tabular}} & $\rho$ & \multicolumn{4}{c|}{1.00} & \multicolumn{4}{c|}{0.84} & \multicolumn{4}{c|}{0.50} \\ \cline{2-14}
		& \begin{tabular}[c]{@{}c@{}}$\alpha$\end{tabular} & 100\% & 72\% & 31\% & 0\% & 100\% & 72\% & 31\% & 0\% & 100\% & 72\% & 31\% & 0\% \\ \hline
		\multicolumn{2}{|c|}{Ours} & 0.549 & 0.581 & 0.558 & 0.569 & 0.685 & 0.698 & 0.686 & 0.693 & 1.037 & 0.974 & 1.080 & 0.925 \\ \hline
		\multicolumn{2}{|c|}{E2E} & 0.602 & 0.660 & 0.864 & - & 0.702 & 0.784 & 0.933 & - & 0.894 & 0.938 & 1.087 & - \\ \hline
		\multicolumn{2}{|c|}{E2E$_{\Delta t}$} & 0.716 & 0.731 & 0.745 & 0.700 & 0.810 & 0.819 & 0.820 & 0.743 & 0.897 & 0.876 & 0.905 & 0.897 \\ \hline
		\multicolumn{2}{|c|}{\begin{tabular}[c]{@{}c@{}}Linear\\ prediction\end{tabular}} & \multicolumn{4}{c|}{2.180} & \multicolumn{4}{c|}{2.162} & \multicolumn{4}{c|}{1.777} \\ \hline
		\multicolumn{2}{|c|}{\begin{tabular}[c]{@{}c@{}}Dataset\\ variance\end{tabular}} & \multicolumn{4}{c|}{0.305} & \multicolumn{4}{c|}{0.296} & \multicolumn{4}{c|}{0.248} \\ \hline
	\end{tabular}
\end{table*}

\subsubsection{Prediction accuracy v.s. scene correlation}
We explore how the correlation $\rho$ between scenes influences our algorithm, which is taking datasets of the same $\alpha$ but different $\rho$ to experiment. We take datasets with $\alpha=31\%$ for example.

\begin{figure}[h]
	\centering
	\includegraphics[keepaspectratio=true,width=\linewidth]{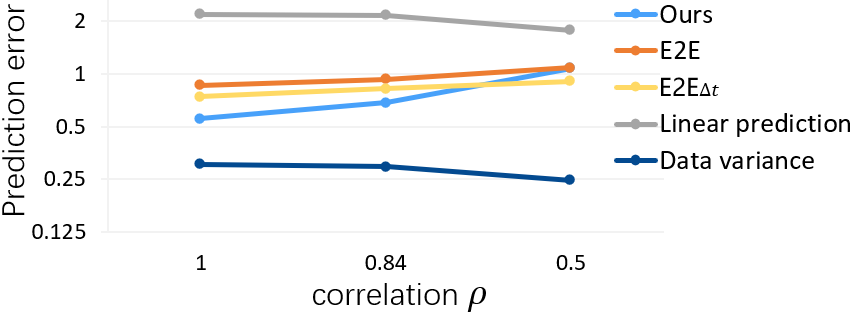}
	\caption{Average prediction error changes with scene correlation level $\rho$, a result of $\alpha$=31\%.  }
	\label{coefficient}
\end{figure}

Quantitative analysis is shown in Fig. \ref{coefficient}. When there is high correlation($\rho=1/0.84$) between scenes, ours(blue) is better than other methods. E2E and E2E$_{\Delta t}$ model have no prior knowledge of scenes but only learn the data mapping of two scenes, and that's why they are worse than ours in high correlation situation. The prediction error of ours increases with the decrease of correlation $\rho$ because the core idea of our method is the latent space of two scenes is shared only when the dynamic change of scenes is correlated. When the correlation between scenes decreases, the performance is down. And that's why when the scenes are less correlative i.e. $\rho=0.5$, the prediction error of ours is larger than E2E/ E2E$_{\Delta t}$ methods. 
The linear prediction model is always the worst.
There are the same results for other $\alpha$ shown in TABLE \ref{predictionAccuracy}.

Qualitative case study is illustrated in Fig.\ref{predictionResults}(a).
Error map $A_1$ is almost white which means our method achieves good result in high correlation situation. From $A_1$ to $A_3$, with the decrease of correlation $\rho$, the error maps become darker and darker, meaning worse and worse prediction results, and our result $A_3$ is even worse than $E2E_{\Delta t}$'s result $C_3$ in less correlation ($\rho=0.5$) situation.

\subsubsection{Prediction accuracy v.s. non-pairwise observation}
We discuss the influence of percentage $\alpha$ of pairwise data, that is taking datasets of the same $\rho$ but different $\alpha$ to experiment. We take datasets with $\rho=0.84$ for example. 

\begin{figure}[htp]
	\centering
	\includegraphics[keepaspectratio=true,width=\linewidth]{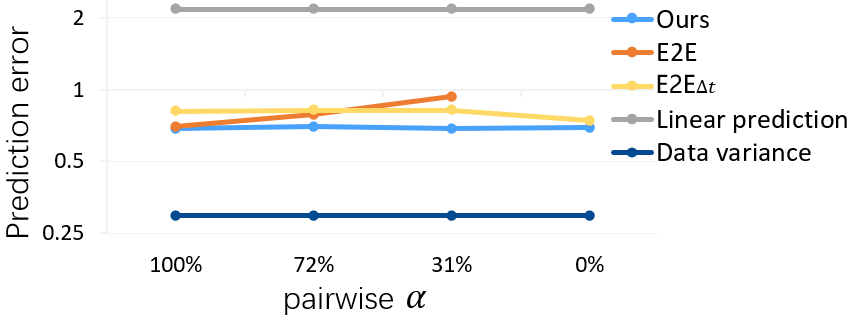}
	\caption{Average prediction error changes with the percentage of pairwise observations $\alpha$, a results of $\rho$=0.84.}
	\label{pairdDataPrecent}
\end{figure}

Quantitative analysis is illustrated in Fig. \ref{pairdDataPrecent}. The prediction error of our method is always the lowest in all percentage $\alpha$. Ours(blue) and E2E$_{\Delta t}$(yellow) method are not sensitive to if scene maps are pairwise, because the time difference between scene maps is considered in them. The E2E method only processes pairwise data in the training step, so the decrease of pairwise data leads to the reduction of training data, causing the prediction error to raise. And that's also the reason for the lack of results on 0\% paired data of E2E method.
There are the same results for other correlation $\rho$ in TABLE \ref{predictionAccuracy}.

Qualitative case study is shown in Fig. \ref{predictionResults}(b). Percentage $\alpha$ does not influence a lot on our methods, and the slight difference between prediction error lead to the similar error maps of all methods.

\subsubsection{Study on single dataset}
\begin{figure*}[]
	\centering
	\includegraphics[keepaspectratio=true,width=\linewidth]{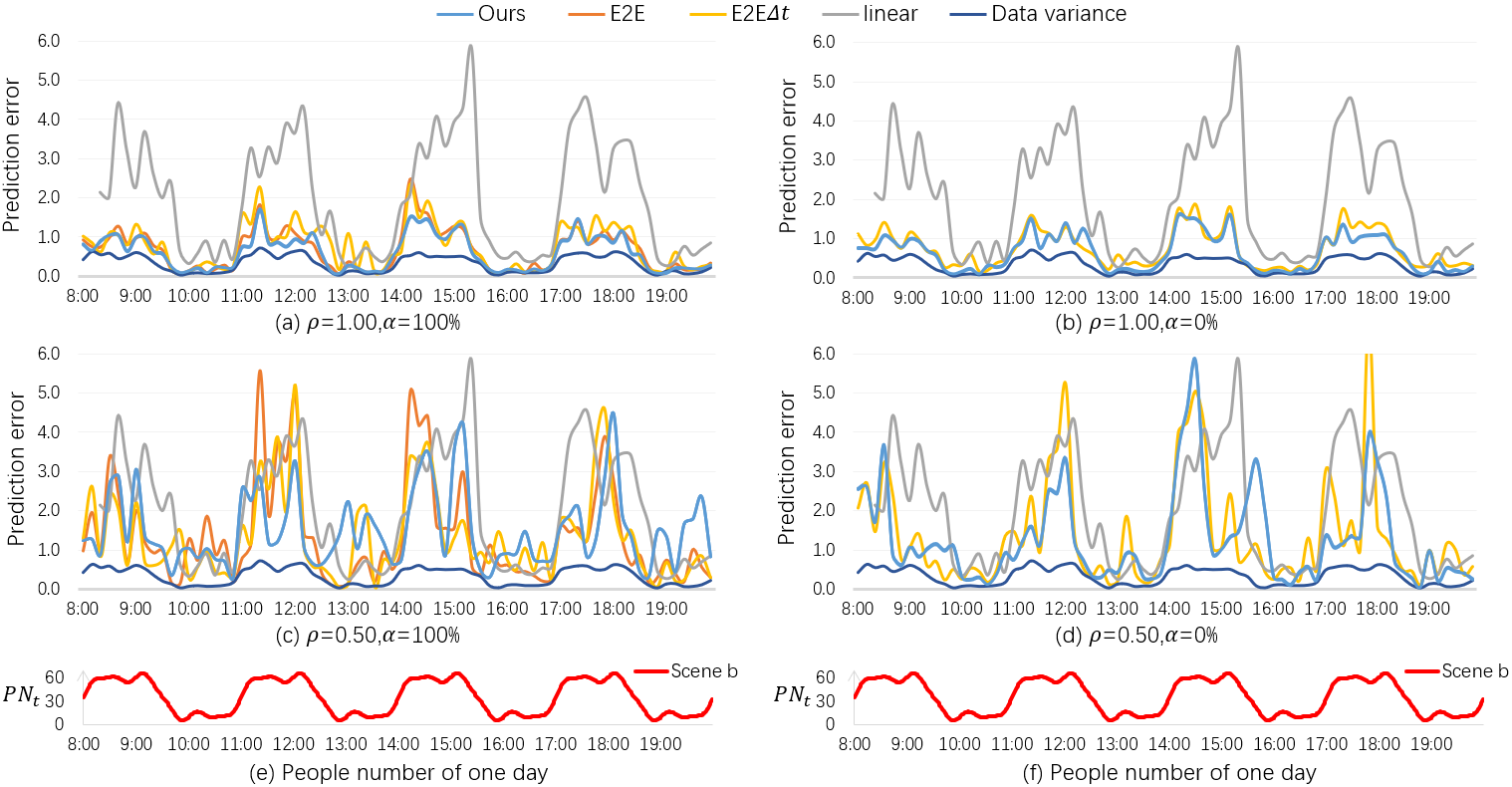}
	\caption{Per map prediction error on each dataset corresponding to a pair of $\rho$ and $\alpha$.}
	\label{fig:studyOnADataset}
\end{figure*}

There are similar results in the study on per map prediction error on single dataset shown in Fig. \ref{fig:studyOnADataset}. Our prediction error is close to data variance  and always lower than baseline methods through the day when the correlation is strong $\rho=1$, and the percentage $\alpha$ of pairwise scene maps rarely influence the performance of our method, shown in Fig. \ref{fig:studyOnADataset}(a)\&Fig. \ref{fig:studyOnADataset}(b). But when there is less correlation $\rho=0.5$ between scenes, ours sometimes can be worse than baseline methods, shown in Fig. \ref{fig:studyOnADataset}(c)\&Fig. \ref{fig:studyOnADataset}(d). 
Finally, Fig. \ref{fig:studyOnADataset}(e) \& Fig. \ref{fig:studyOnADataset}(f) are the people number of one day, and the data variance changes with it. This is because in our pedestrian simulator, every pedestrian's movement is influenced by its nearby people, and when there are lots of people in the scene, the randomness of pedestrians' movement increases, leading to the raising of data variance.

\section{CONCLUSIONS}
This paper is the first try to answer the question: can we make inference by modeling the correlations of scene dynamics on history observations?
We formulate the problem as given a set of unsynchronized history observations of two scenes that are correlative on their dynamic changes, learn a cross-scene predictor, wherewith the observation of one scene, a robot can onlinely predict the dynamic state of another.
The problem is solved by developing a method by modeling the inherent correlation of scene dynamics using latent space shared auto-encoders,
where a learning model is established by connecting two auto-encoders through the latent space, and a prediction model is built by concatenating the encoder of the input scene with the decoder of the target one. 
The method is examined through simulation, where the dynamic flows at two adjacent gates of campus are imitated.
The problem is adaptive to other scenarios such as successive intersections on a single road, gates of subway stations, etc., where the dynamic changes are triggered some common events.
Cross-scene prediction accuracy is examined at various conditions of scene correlation and pairwise observations, 
and the results show that the proposed method can better solve the problem than the conventional end-to-end and linear predictions ones.
Future work will be addressed on real-data collection and processing, and the inference on dynamic correlations of more adjacent scenes will also be studied.

\bibliographystyle{IEEEtran}
\bibliography{ref}
\end{document}